%% file: main.tex
\documentclass[twoside]{article}
\usepackage{aistats2014}

\usepackage{hyperref}
\usepackage{url}
\usepackage{sidecap}
\usepackage{floatrow}

\usepackage{amsmath,amssymb,bbm}
\usepackage[linesnumbered,ruled,vlined]{algorithm2e}
\DeclareCaptionType{copyrightbox}
\usepackage{subfig}
\usepackage{graphicx}
\usepackage{color}
\usepackage{bm}
\usepackage[numbers]{natbib}
\usepackage{IEEEtrantools}
\usepackage{anyfontsize}
\newcommand\ie {{\it i.e., \ }}
\newcommand\eg {{\it e.g., }}

\newcommand\etal {{\it et al.}}
\newcommand\eq {{\it Eq.}}
\newcommand\boldTheta{{\bm{\theta}}}
\newcommand\boldGamma{{\bm{\gamma}}}
\newcommand\boldPhi{{\bm{\Phi}}}

\begin{document}



\twocolumn[

\aistatstitle{Diversifying Sparsity Using Variational Determinantal Point Processes}

%\aistatsauthor{ Anonymous Author 1 \And Anonymous Author 2 \And Anonymous Author 3 }
\aistatsauthor{ N.K. Batmanghelich$^1$ \And G. Quon$^1$  \And A. Kulesza$^2$  
\And M. Kellis$^1$ \And P. Golland$^1$ \And L. Bornn$^3$ }

%\aistatsaddress{ Unknown Institution 1 \And Unknown Institution 2 \And Unknown Institution 3 } 
\aistatsaddress{$^1$Massachusetts Institute of Technology \And $^2$University of Michigan \And $^3$Harvard University } 
]

\begin{abstract}
  We propose a novel diverse feature selection method based on determinantal point processes (DPPs). Our model enables one to flexibly define diversity based on the covariance of features (similar to orthogonal matching pursuit) or alternatively based on side information. We introduce our approach in the context of Bayesian sparse regression, employing a DPP as a variational approximation to the true spike and slab posterior distribution. We subsequently show how this variational DPP approximation generalizes and extends mean-field approximation, and can be learned efficiently by exploiting the fast sampling properties of DPPs. Our motivating application comes from bioinformatics, where we aim to identify a diverse set of genes whose expression profiles predict a tumor type where the diversity is defined with respect to a gene-gene interaction network. We also explore an application in spatial statistics. In both cases, we demonstrate that the proposed method yields significantly more diverse feature sets than classic sparse methods, without compromising accuracy.
\end{abstract}

\input{introduction_v2.tex}

\input{method_v2.tex}

\input{experiments.tex}

\input{conclusion.tex}

\appendix
\input{appendix.tex}

{%\fontsize{9}{10}\selectfont 
\small
\bibliographystyle{plainnat}
\bibliography{mybib}
}

\end{document}

%% file: introduction_v2.tex
\section{Introduction}

As modern technology enables us to capture increasingly large amounts of data, it is critically important to find efficient ways to create compact, functional, and interpretable representations. Feature selection is a promising approach, since reducing the feature space both improves interpretability and prevents over-fitting; as a result, it has received considerable attention in the literature \citep[e.g.,][]{tibshirani1996regression, guyon2003introduction}. In this paper, we focus on the problem of \emph{diverse} feature selection, where the notion of diversity can be defined in terms of the features themselves or in terms of available side information.

%Diverse feature sets have the potential to be both more compact and easier to interpret, without sacrificing performance.  Diversity also plays a more fundamental role in some real-world applications; for example, cancers can be very heterogeneous \cite{garcia2008heterogeneity} and require identification of diverse biomarkers for the prognosis of different tumor subtypes.  Thus, we aim to identify a diverse set of genes whose expression profiles collectively predict a tumor type, where the diversity is defined with respect to a separate gene-gene interaction network (see Section \ref{sec:genNet}).

Diverse feature sets have the potential to be both more compact and easier to interpret, without sacrificing performance.  Diversity also plays a more fundamental role in some real-world applications; for example, breast cancer is increasingly recognized to present a highly heterogeneous group of malignancies~\cite{garcia2008heterogeneity} where subgroups may involve different mechanisms of action. For the common task of identifying gene expression-based biomarkers of different tumor subtypes, maximizing the diversity of selected genes helps identifying these disparate mechanisms of action. Diversity in this case is defined with respect to a separate gene-gene interaction network (see Section \ref{sec:genNet}).

Existing techniques for feature selection generally do not explicitly consider feature diversity.  From an optimization point of view, feature selection can be viewed as a search over all possible subsets of features to identify the optimal subset according to a pre-specified metric, often balancing model fit and model complexity.  To avoid enumerating the entire combinatorial search space, embedded approaches, such as the LASSO \citep{tibshirani1996regression}, relax the problem to a combination of a sparsity term ($\ell_1$) and a data fidelity term. However, such methods typically do not encourage diversity explicitly. In fact, LASSO has been shown to be unstable in the face of nearly collinear features \citep{grave2011trace}, with several variants proposed to ameliorate this issue \citep{Zou:2005aa}.  

An alternative approach is to search greedily by successively adding (or removing) the best (or worst) feature~\citep{guyon2003introduction}. Orthogonal matching pursuit (OMP) proceeds in this way using forward step-wise feature selection, but the selected feature is chosen to be as orthogonal as possible to previously selected features \citep{pati1993orthogonal}. One can view this orthogonality as a measure to implicitly maximize diversity \cite{das2011submodular}. In spite of its well-established performance \citep{tropp2007signal}, OMP is a procedure that lacks an underlying generative model and, therefore, the flexibility to define diversity other than through the inner product of features.

In this paper we take a probabilistic view of the problem, assigning a probability measure to feature subsets. We then seek the maximum \textit{a posteriori} (MAP) estimate as the optimal subset.  In particular, our probability measure is a variational approximation to the posterior of the spike-and-slab variable selection model.  By imposing a particular form on that approximation, we obtain a measure that assigns higher scores to feature subsets that are not only relevant to a regression or classification task but also diverse.  The challenge is to find a form that achieves this goal while remaining computationally tractable.

To this end, our variational approximation takes the form of a determinantal point process (DPP).  DPPs are appealing in this context since they naturally encourage diversity, defined in terms of a kernel matrix that can be, for example, the feature covariance matrix (discouraging collinearity), or alternatively derived from application-specific notions of similarity~\cite{kulesza2012determinantal}.  DPPs also offer computationally appealing properties such as efficient sampling and approximate MAP estimation \cite{mapDPP,nemhauser1978analysis}.  As a result, not only can we efficiently approximate the optimal feature set, but we can also provide sampling-based credible intervals.

Unlike mean field approximations that fully factorize the posterior distribution, our approximate DPP posterior has a complex dependency structure.  This makes fitting the DPP a challenging task.  Kulesza \etal \cite{kulesza2012determinantal} used an optimization approach to learn  conditional DPPs, and Affandi \etal \cite{affandi_learningDPP} proposed to parameterize the kernel of DPPs and learn the parameters using a sampling approach; however, neither approach allows learning a full, unparameterized kernel.  In contrast, our algorithm is based on a flexible variational framework proposed by Salimans and Knowles \cite{fixFormVBLinReg} that only requires two basic operations: efficient evaluation of the joint likelihood and efficient sampling from the current estimate of the posterior.  Fortunately, for DPPs these operations are efficient.
% not sure what these commented sentences mean (--Alex):
%; further, our approach is applicable when the marginal likelihood of the collapsed model can be approximated. 
%(\ie marginal likelihood of the model marginalizing the coefficients conditioned on the subset), 
For regression, the marginal likelihood takes a closed form, and for many other models, including classification, it can be approximated efficiently. 
%, (\eg using expectation propagation \cite{GPBook}). 
To the best of our knowledge this work presents the first use of DPPs within a variational framework.

The closest work to ours is by George \etal\cite{george2010dilution}. They suggested variations of so-called \emph{dilution} priors that, instead of assigning prior uniformly across models, assign probability more uniformly across \textit{neighborhoods}, and then distribute the neighborhood probabilities across the models within them.  One of the diluting priors proposed in \cite{george2010dilution} is proportional to the determinant of the features, resembling the form of a DPP. Such a prior does not guarantee diversifying properties on the posteriori selected features. Although possible, instead of prior, we suggest to approximate the posterior with DPP. We investigate this model choice in Section \ref{sec:experiment}.

This work makes the following contributions.  We propose to use a DPP as an approximate posterior for Bayesian feature selection. To fit the variational approximation, we draw a connection between DPPs and the exponential family using the decomposition proposed by Kulesza \etal \cite{kulesza2012determinantal}. This connection makes many tools developed for the exponential family available for DPPs, including variational learning methods for distributions that are not fully factorizable. Our proposed method brings a number of advantages, including the ability to: (i) propose multiple sets of relevant and diverse features which can be viewed as alternative feature selection solutions; (ii) characterize feature selection uncertainty through posterior sampling; (iii) flexibly define feature set diversity based on side information, rather than just covariance; and (iv) compute the marginal probability for inclusion of new features conditioned on the presence of an existing feature set, thanks to the computational properties of DPPs.

The remainder of the paper is organized as follows.  We first review DPPs in Section \ref{sec:bkgDPP}. The idea of diverse sparsity is illustrated in the context of Bayesian variable selection in Section \ref{sec:model}. In Section \ref{sec:learning}, we show how to learn the parameters of such models. In the Section \ref{sec:experiment}, we apply the method to identify a diverse set of genes to predict a tumor type while the diversity is defined with respect to a gene-gene interaction network. Finally, in Section \ref{sec:optGrid}, we explore application to learning an optimal distribution of grid points in a spatial process convolution model.
%A third application to imaging genetics is described in the supplementary materials, where the goal is to select an uncorrelated subset of brain regions that are relevant to clinical measurements and genetic observations.

%% file: method_v2.tex
\section{Determinantal Point Process}
\label{sec:bkgDPP}
The determinantal point process (DPP) defines distribution over configurations of points in space.  
%It provides a distribution over subsets of space. 
If the space is finite, say $[M]:=\{1,\cdots,M \}$; it defines probability mass over all $2^M$ subsets. Specifically, the probability of a subset $\boldGamma \in \{0,1 \}^M$ is proportional to the determinant of  $\left[ \mathbf{L} \right]_{\bm{\gamma}}$ where $\left[ \cdot \right]_{\bm{\gamma}}$ denotes the submatrix containing the columns and rows $i$ for which $\gamma_i = 1$  and $\mathbf{L} \in \mathbb{R}^{M\times M}$ is a positive semidefinite kernel matrix. Strictly speaking, this representation defines a subclass of DPPs called $L$-ensembles.  If $L_{ij}$ is a measurement of similarity between elements $i$ and $j$, then the DPP assigns higher probabilities to subsets that are diverse.  More precisely, if $L_{ij} = \phi(i)^T \phi(j)$ for a feature function $\phi : [M] \to \mathbb{R}^d$, then the probability of a set $\boldGamma$ is proportional to the squared volume spanned by $\{\phi(i)\ |\ \gamma_i = 1 \}$.  Elements with orthogonal feature functions will tend to span larger volumes, and hence have a higher probability of co-occurrence. 
%DPPs have computationally appealing properties, including efficient marginalization and sampling. 
For more in-depth review of DPP and its applications in machine learning see \cite{kulesza2012determinantal}. 

\begin{figure}[tb]
	\centering
	\subfloat[][]
	{
		%\begin{minipage}{5.0cm}
		\includegraphics[height=3.2cm]{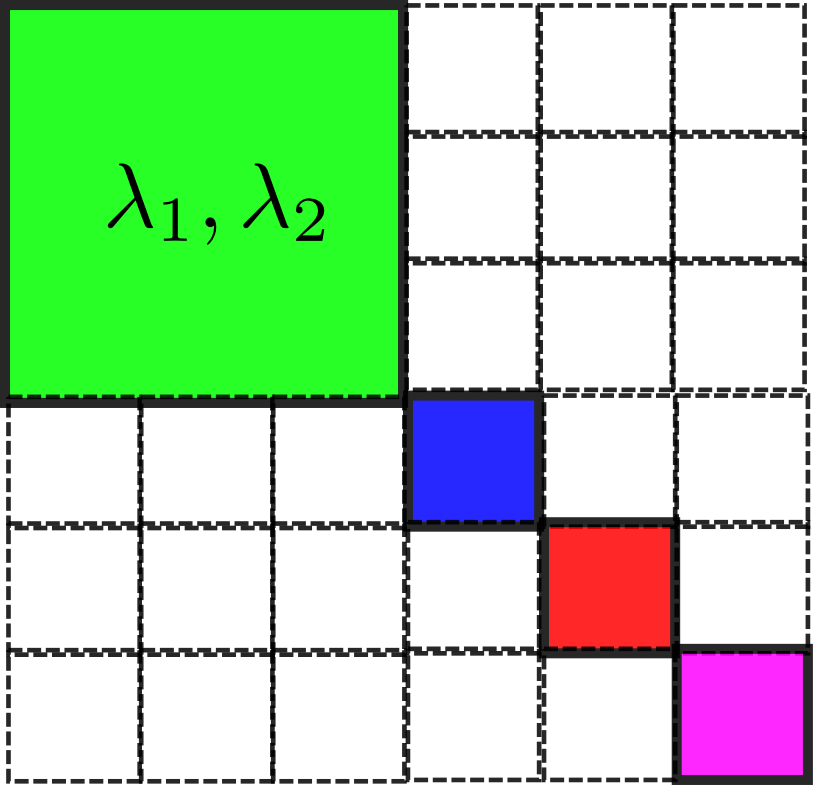}
		%\end{minipage} 
		\label{fig:lowRankCov}
	} 
        \subfloat[][]
	{
	      \includegraphics[height=3.2cm]{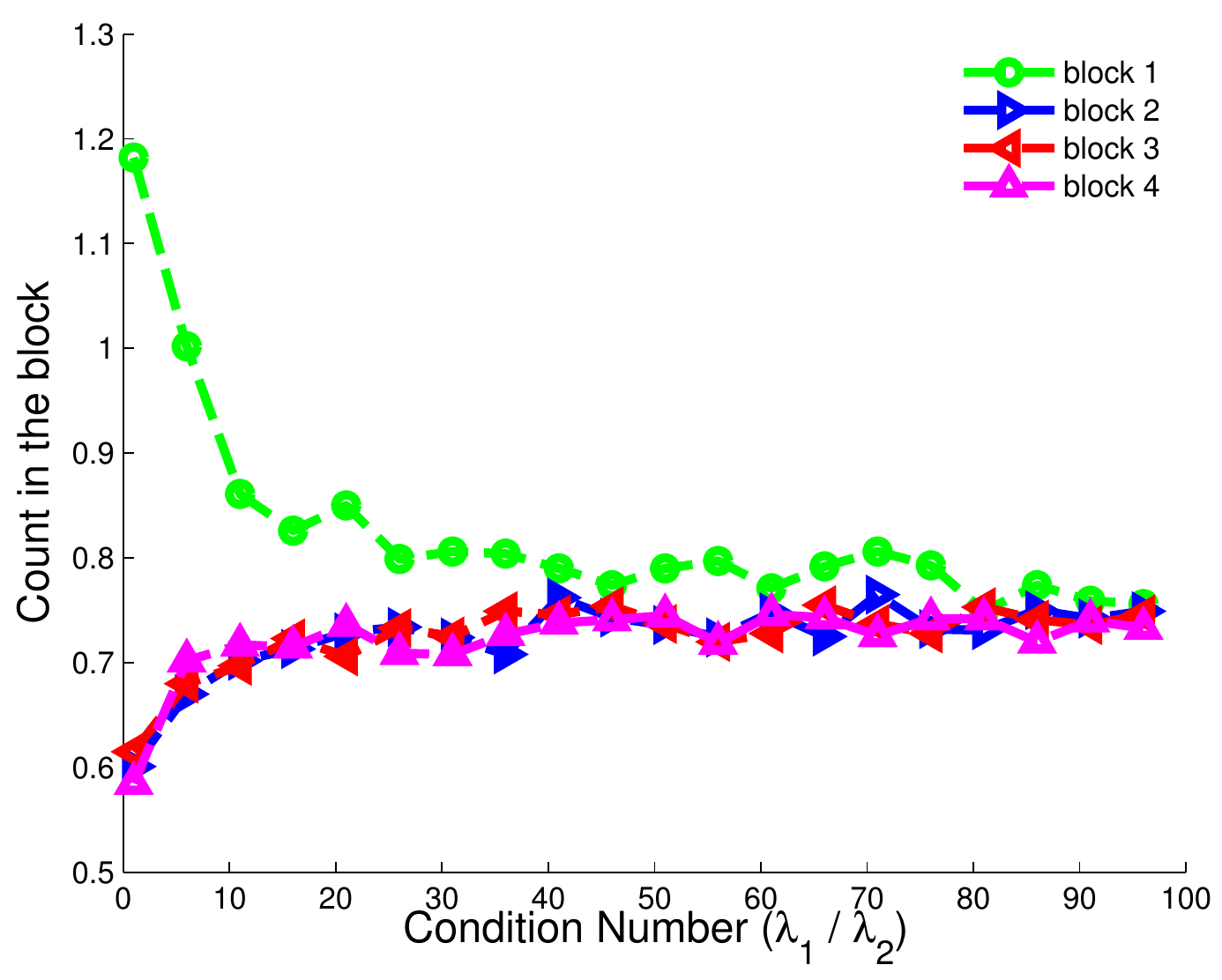}
	      \label{fig:covSampling}
	}
	\caption{\protect\subref{fig:lowRankCov} $L$-ensemble of DPP for a toy problem of six items. The items are dissimilar except the first three items. This leads to a block diagonal kernel  $\mathbf{L}$. \protect\subref{fig:covSampling} Empirical average number of elements selected from each block. Although the green block is bigger than the other blocks, as it becomes more collinear $( \frac{\lambda_1}{\lambda_2} \rightarrow \infty )$, the probability of selecting an item from the first block converges to that of the other blocks.}
	\label{fig:lowRankCovDPPSampling}
\end{figure}

In addition to having computationally appealing properties such as efficient marginalization and sampling, repulsive interactions are also modeled elegantly through a DPP. To illustrate this preference for diversity, suppose we would like to choose a subset from six items.  The items are dissimilar except for the first three items. Therefore, their $L$-ensemble matrix is block diagonal where the first three items form a single group, illustrated as a green block in Figure \ref{fig:lowRankCov}. We assume that the green block has rank two with eigenvalues of $\lambda_1$ and $\lambda_2$. As the condition number $\frac{\lambda_1}{\lambda_2}$ increases, the items of the green block become more similar (collinear). We can sample from this DPP and compute the empirical average of samples falling into each block as a function of $\frac{\lambda_1}{\lambda_2}$ (Figure \ref{fig:covSampling}). If there is no interaction, the probabilities are proportional to the sizes of the blocks but as $\frac{\lambda_1}{\lambda_2} \rightarrow \infty$, the probability of selecting an item from the first block 
decreases to that of the rest of the blocks.

\section{Methods}
\subsection{Bayesian Variable Selection}
\label{sec:model}
 Following standard notation, we consider the regression model $\mathbf{y} = \sum_{m=1}^{M} \mathbf{x}_m \beta_m + \varepsilon$, where the regressors $\{ \mathbf{x}_1 , \cdots, \mathbf{x}_M \}$ are collected into a design matrix $\mathbf{X} = \left[ \mathbf{x}_1 , \cdots , \mathbf{x}_M \right] \in \mathbb{R}^{N \times M}$ and $\varepsilon$ is the residual noise $\varepsilon \sim \mathcal{N}(\cdot; 0,\sigma)$. One can view variable selection as deciding which of the coefficients  $\beta_m$ are nonzero. This is often made explicit in Bayesian variable selection through a latent binary random vector $\bm{\gamma} \in \{0,1 \}^{M}$ that specifies which predictors are included: 
\begin{equation}
  \mathbf{y} =  \mathbf{X} ( \bm{\gamma} \odot \bm{\beta} ) + \varepsilon,
\end{equation}
where $\odot$ is the element-wise product. Assuming an exchangeable Bernoulli prior for $\boldGamma$ and a conjugate prior for $\bm{\beta}$ with covariance $\sigma^2 \Lambda_0^{-1}$, \ie $\bm{\beta} \sim \mathcal{N}(\cdot; 0,  \sigma^2 \Lambda_0^{-1} )$, random variable $\gamma_m \beta_m$ defines the so-called ``spike-and-slab'' prior \cite{carbonetto2012scalable, mitchell1988bayesian}.  It is drawn from the spike with probability $\alpha$ and from the slab with probability $1-\alpha$. Assuming an inverse Gamma prior with parameters $(a_0,b_0)$ for variance $\sigma^2 $, the joint likelihood of the model can be written as follows:
\begin{IEEEeqnarray}{ll} 
  p (\mathbf{y}, \pi ; \mathbf{X}, \rho) = &   \\ 
   & p (\mathbf{y}|  \bm{\gamma}, \bm{\beta}, \sigma ; \mathbf{X})  p (\bm{\beta}| \sigma; \Lambda_0) p(\sigma; a_0, b_0) p ( \boldGamma ; \alpha),   \nonumber 
  \label{eqn:spike-Slab}
\end{IEEEeqnarray}
where $\pi = \{ \bm{\beta},\sigma,\boldGamma \}$ is the set of latent random variables and $\rho = \{ \Lambda_0, a_0 , b_0, \alpha \}$ is the set of fixed parameters. We set $\Lambda_0 = c \mathbf{I}$.

Conditioned on  $\bm{\gamma}$, the marginal likelihood of the restricted regression can be expressed in closed-form \cite{carlin2008bayes}:
\begin{IEEEeqnarray}{l}
   \log{ p \left( \mathbf{y} | \bm{\gamma} ; \mathbf{X} , \rho \right)} =   \\ 
   -\dfrac{N}{2} \log{2 \pi} + 
    \dfrac{1}{2} \left(\log{\det \left( \Lambda_0 \right)} -\log{ \det \left( \Lambda_N \right)} \right) +  \nonumber \\
    (a_0 - a_N) \left( \log b_0 - \log b_N \right)   +   \log{\Gamma \left( a_N \right)} - \log{\Gamma \left( a_0 \right)}  \nonumber \\
   \Lambda_N =  \mathbf{X}^T \mathbf{X} + \Lambda_0,  \quad \quad   \bm{\mu}_N =  \Lambda_N^{-1} \mathbf{X}^T \mathbf{y}   \nonumber 
   \label{eqn:margLikli}
\end{IEEEeqnarray}
where $\Gamma \left( \cdot \right)$ is the Gamma function,  $a_N =  a_0 + N / 2$, and $b_N =  b_0 + \frac{1}{2}  \left( \mathbf{y}^T \mathbf{y} - \bm{\mu}_N^T \Lambda_N \right)$.

Exact inference of the posterior inclusion probability of the regressors, \ie $p ( \boldGamma | \mathbf{y} ; \mathbf{X}, \rho)$, is computationally prohibitive since it entails a sum over all possible subsets of $[M] := \{1,\cdots,M \}$. We therefore resort to an approximation. 
Variational approaches approximate the form of the posterior; for example, the mean field approach employs a fully factorized function as the approximating distribution~\cite{beal2003, carbonetto2012scalable}. Marginalizing $\bm{\beta}$ out, the mean field approximates the posterior probability of variable inclusion as
\begin{IEEEeqnarray}{l}
  q (\bm{\gamma}; \bm{\theta}) = \prod_{m = 1}^M q (\gamma_m ; \theta_m ) = \prod_{m = 1}^M \theta_m^{\gamma_m} (1 - \theta_m)^{1 - \gamma_m}
  \label{eqn:post-meanField}
\end{IEEEeqnarray}
However, this form of posterior does not account for interaction between regressors, diversity being one such form. 
%For example, one can view orthogonality between predictors as an instance of diversity (as in orthogonal matching pursuit). In some applications (\eg bioinformatics), one may prefer to choose a diverse yet relevant set of variables (\eg genes) that explain the response $y$ (\eg phenotype) well. For example, if expression levels of two genes are very correlated, or they reside on a common pathway, we would like to choose an alternative subset of dissimilar genes to explain $y$. 

To encourage diversity, one needs to model the interactions between features. We propose to use a DPP in an elegant way to define probability mass over all possible subsets of $[M]$. As a naming convention, ``\texttt{XX-YY}'' refers to prior specification \texttt{XX} and variational distribution \texttt{YY}; hence we will refer to the setting as \texttt{Bernoulli-DPP}. It is possible to define the DPP as a prior for $\bm{\gamma}$ and approximate the posterior with a fully factorized mean field method, \ie Bernoulli (referred as \texttt{DPP-Bernoulli}). However, such a prior does not guarantee that the posterior exhibits the diversifying property. It is also straightforward to have DPP as prior and posterior (\ie \texttt{DPP-DPP}) but here we focus on investigating how effective DPP is as prior versus posterior.

Following the formulation of Kulesza \etal \cite{kulesza2012determinantal}, we propose the following variational posterior distribution:
\begin{IEEEeqnarray}{ll}
  q (\bm{\gamma}; \bm{\theta} ) &= \dfrac{1}{Z_{\bm{\theta}}} \det \left[ \mathbf{L} \right]_{\boldGamma}  
                                = \dfrac{1}{Z_{\bm{\theta}}} \det  \left[ \text{diag}(  e^{ \frac{\bm{\theta}}{2} }  )  \bm{\Phi} \bm{\Phi}^T  \text{diag}(  e^{ \frac{\bm{\theta}}{2} }  )   \right]_{\bm{\gamma}} \nonumber \\
                                 &= \dfrac{1}{Z_{\bm{\theta}}} e^{ \bm{\theta}^{T}  \boldGamma  } \det  \left[   \bm{\Phi} \bm{\Phi}^T \right]_{\bm{\gamma}} 
  \label{eqn:post-dpp}
\end{IEEEeqnarray}
where $\boldTheta$ and $\boldGamma$ are parameters and latent random variables respectively, and  $Z_{\bm{\theta}} = \det( \mathbf{I} + \mathbf{L} )$ is the normalization constant \cite{kulesza2012determinantal}. $\bm{\Phi} \in \mathbb{R}^{M \times d} $ is a given matrix of similarity features whose row $m$,  $\phi(m)$, is the similarity feature vector for item $m$, and $L$-ensemble matrix is $\mathbf{L} = \text{diag}(  e^{ \frac{\bm{\theta}}{2} }  )  \bm{\Phi} \bm{\Phi}^T  \text{diag}(  e^{ \frac{\bm{\theta}}{2} }  ) $. For example if $\bm{\Phi} = \mathbf{X}$, the DPP discourages collinearity. $\bm{\Phi}$ can also be defined via side information (see Section \ref{sec:experiment}).

Note that \eq(\ref{eqn:post-dpp}) reduces to \eq(\ref{eqn:post-meanField}) if the similarity features are indicator vectors, \ie $ \bm{\Phi} \bm{\Phi}^T  = \mathbf{I}$. In this case, the DPP approximation proposed here reverts to a mean field approximation. 

\subsection{Learning}
\label{sec:learning}
We now propose an algorithm to fit the variational approximation for both \texttt{DPP-Bernoulli} and \texttt{Bernoulli-DPP}. Learning with \texttt{Bernoulli-DPP} is more challenging than \texttt{DPP-Bernoulli}. To see why, note that the variational approach minimizes the divergence
\begin{IEEEeqnarray}{l}
  \text{KL} \left( q_{\boldTheta} | p (\boldGamma, y) \right)= \mathbbm{E}_{q_{\boldTheta}}  \left[ \log q_{\boldTheta} \left( \boldGamma \right) -  \log p \left( \boldGamma, y \right) \right]
\end{IEEEeqnarray}

When $q_{\boldTheta}$ is a DPP, computing the first term, which is the entropy of the DPP entails  $2^M$ summands which, to the best of our knowledge, does not have any closed-form. We focus on approximating \texttt{Bernoulli-DPP} and show how a straightforward modification to the resulting algorithm can effectively approximate the posterior of  \texttt{DPP-Bernoulli}. 

%The problem of learning DPP parameters from data remains open. Kulesza \etal \cite{learnDPP} proposed a maximum likelihood approach to learn the parameters of conditional DPPs and showed that it leads to a convex and efficient learning formulation. Affandi \etal \cite{affandi_learningDPP} proposed to parameterize the kernel of DPPs and learn the parameters using a sampling approach. In contrast, our algorithm is based on a flexible variational framework proposed by Salimans and Knowles \cite{fixFormVBLinReg} that only demands efficient sampling from the posterior and closed form joint likelihood. Fortunately, sampling from a DPP is efficient; further, for models where the restricted marginal likelihood can be approximated (\ie conditioned on the subset), this model is applicable. For regression, the marginal quantity takes the closed-form of \eq\ref{eqn:margLikli} and for many other models including classification, it can be approximated (\eg using Expectation Propagation \cite{GPBook}). To the best of our 
%knowledge this is the first variational approach to learning DPPs.
To learn $\boldTheta$ in \eq(\ref{eqn:post-dpp}), we borrow the stochastic approximation algorithm from \cite{fixFormVBLinReg}, which allows one to approximate any distribution that is given in a closed-form. In \emph{structured} or \emph{fixed-form} variational Bayes \cite{honkela2010approximate}, the posterior distribution is chosen to be a specific member of an exponential family, namely $q (\boldGamma ; \boldTheta) = \exp \left( \boldTheta^T T(\boldGamma) -U  \left( \boldTheta \right) \right) \nu \left( \boldGamma \right)$ where $T(\boldGamma)$ is the sufficient statistic, $U(\boldTheta)$ is the normalizer and $\nu(\boldTheta)$ is a base measure. The DPP in its general form is not a member of exponential family but parameterizing DPPs as \eq (\ref{eqn:post-dpp}) allows us to employ the framework. We first summarize the algorithm in \cite{fixFormVBLinReg} in  our context where $T(\boldGamma):=\boldGamma$, $\nu(\boldGamma):= \det ( [ \bm{\Phi} \bm{\Phi}^{T} ]_{\boldGamma})$, and $U( \boldTheta  ) := \det ( \mathbf{L} + \mathbf{I} )$. 

For notational convenience, we define $\tilde{q}_{\tilde{\boldTheta}} : = \exp ( \tilde{\boldGamma}^{T}  \tilde{\boldTheta} ) $ where $\tilde{\boldTheta}^{T} = \left[ \boldTheta^T , \theta_0 \right]$ 	and $\tilde{\boldGamma}^{T} = \left[ \boldGamma^T , 1 \right] $. If $\theta_0 = -U(\boldTheta)$, then $\tilde{q}$ is the normalized posterior, otherwise it is an unnormalized version \cite{fixFormVBLinReg}.
Taking the gradient of the unnormalized KL-divergence $\mathcal{D} (\tilde{q}_{\tilde{\boldTheta}} | p(\boldGamma,\mathbf{y}))$, with respect to $\tilde{\boldTheta}$ we obtain:
\begin{IEEEeqnarray}{ll}
  \nabla_{\tilde{\boldTheta}} \mathcal{D}  & \left[ q_{\tilde{\boldTheta}} | p \left( \boldGamma, y \right) \right] =  \nabla_{\tilde{\boldTheta}}  \mathbbm{E}_{\tilde{q}}  \left[ \log \tilde{q}_{\tilde{\boldTheta}} \left( \boldGamma \right) -  \log p \left( \boldGamma, y \right) \right] \nonumber   \\
  &= \int \tilde{q}_{\tilde{\boldTheta}}(\boldGamma) \left[ \tilde{\boldGamma} { \tilde{\boldGamma} }^T \tilde{\boldTheta} - \tilde{\boldGamma} \log p( \boldGamma, y  )  \right] d \nu(\boldGamma)  
  \label{eqn:gradKL}
\end{IEEEeqnarray}

By setting \eq(\ref{eqn:gradKL}) to zero, Salimans and Knowles \cite{fixFormVBLinReg} linked linear regression and the variational Bayes method. Namely, at the optimal solution, $\tilde{\boldTheta}$ should satisfy the linear system : $\mathbf{C} \tilde{\boldTheta} = \mathbf{g}$ where  $\mathbf{C} := \mathbbm{E}_q  \left[ \tilde{ \boldGamma } \tilde{ \boldGamma }^T \right]$ and $\mathbf{g} := \mathbbm{E}_q  \left[ \tilde{ \boldGamma } \log p \left( \boldGamma, y \right) \right]$. $\mathbf{C}$ and $\mathbf{g}$ are estimated via weighted Monte Carlo sampling by drawing a single sample, $\boldGamma_t$, from the current posterior
approximation $q_{\tilde{\boldTheta}_t}$,
\begin{IEEEeqnarray}{cl}
  %\mathbf{g}_{t + 1} &= \left( 1 - w \right) \mathbf{g}_t + w \hat{\mathbf{g}}_t   \nonumber\\
  %\mathbf{C}_{t + 1} &= \left( 1 - w \right) \mathbf{C}_t + w \hat{\mathbf{C}}_t  
  \mathbf{g}_{t + 1} &= \left( 1 - w \right) \mathbf{g}_t + w \tilde{ \boldGamma_t } \log p \left( \boldGamma_t, y \right)   \nonumber\\
  \mathbf{C}_{t + 1} &= \left( 1 - w \right) \mathbf{C}_t + w \tilde{\boldGamma_t }  \tilde{ \boldGamma_t }^T 
  \label{eqn:MCUpdate}
\end{IEEEeqnarray}
%where $\widehat{\mathbf{g}}_t = \tilde{ \boldGamma_t } \log p \left( \boldGamma_t, y \right)$, $\widehat{\mathbf{C}}_t = \tilde{\boldGamma_t }  \tilde{ \boldGamma_t }^T$, and $w \in \left[ 0, 1\right]$ is the step size. 
where $w \in \left[ 0, 1\right]$ is the step size.

Interestingly,  $\mathbb{E}_{q} \left[  \boldGamma  \boldGamma ^T \right]$ is the DPP marginal kernel, $\mathbf{K}$, which has the closed form $\mathbf{K}~=~\mathbf{L}(\mathbf{L} + \mathbf{I})^{-1}$. Nevertheless, in our experiments, we did not see any clear advantage in substituting the current estimate for $\mathbf{K}$ directly into \eq(\ref{eqn:MCUpdate}) versus using the empirical estimate.

Pseudo-code for our algorithm is shown in Algorithm~\ref{algo:fixFrmDPP} in Appendix \ref{sec:appendix}.  We set $p(\boldGamma) = \prod_m  \alpha^{\gamma_m} (1 - \alpha)^{1-\gamma_m} $, and as suggested in \cite{fixFormVBLinReg}, $w := \frac{1}{\sqrt{N}}$. We further set the initial $L$-ensemble to $\mathbf{L}~=~(e^{\theta_0/2} \bm{\Phi})( \bm{\Phi}^T e^{\theta_0/2})$, where $\theta_0$ is adjusted to make sure that the initial samples from the DPP are not empty sets. To do so, we note that the diagonal elements of $\mathbf{K}$ are the marginal probability of inclusion of element $i$. Therefore the expected cardinality of the subset is $tr(\mathbf{K})= \sum_i \frac{ e^{\theta_0} \lambda_i }{1 + e^{\theta_0} \lambda_i }$, where $\lambda_i$ is the $i$'th eigenvector of $\bm{\Phi} \bm{\Phi}^T$. To set the expected cardinality of subsets to a preset value $\kappa$, we solve the equation for $\theta_0$.
%we can numerically find $\theta_0$ such that $\sum_{i} \frac{e^{\theta_0} \lambda_i }{1 + \lambda_i e^{\theta_0}} = \kappa$.

 Algorithm~\ref{algo:fixFrmDPP} only requires sampling from a DPP with parameters $\boldTheta_t$ and computing $p(y,\boldGamma_t)~=~p(y|\boldGamma_t)p(\boldGamma_t)$. For example, in the regression problem \eq(\ref{eqn:spike-Slab}), $p(y|\boldGamma_t)$ has the closed-form solution in \eq(\ref{eqn:margLikli}). In a linear logistic regression case (classification),  $p(y|\boldGamma_t)$ does not have a closed-form but conditioned on $\boldGamma_t$, its computation is the equivalent of computing the marginal likelihood for linear kernel Gaussian Process model, which can be approximated using expectation propagation (EP).% \cite{GPBook}. 
  Joint distribution $p(y,\boldGamma_t)$ can encode more involved models as long as $p(y|\boldGamma_t)$ can be approximated efficiently (see the supplementary material for an example). One side benefit of the algorithm is the automated selection of the number of features included in the model. If fixed model size is desired, the algorithm can be easily extended to employ $k-$DPPs, where the cardinality of the subset is 
fixed, by replacing the sampling part of the algorithm. 
 
After learning the DPP, we compute the MAP estimate using \cite{mapDPP, nemhauser1978analysis} to find the most relevant and diverse set.
Other than MAP, we can easily compute a credible interval of our approximation of $y$ by drawing samples from the approximate posterior, predicting $y$ for each draw, and computing the variance of the prediction.

In Algorithm \ref{algo:fixFrmDPP}, we focused on having Bernoulli prior and DPP posterior, \ie \texttt{Bernoulli-DPP}. To adapt the algorithm for \texttt{DPP-Bernoulli}, we modify $p(\mathbf{y},\bm{\gamma})$ by changing the prior to $p(\bm{\gamma}) = \frac{\det ( [\mathbf{L}]_{\bm{\gamma}} )}{\det ( \mathbf{L} + \mathbf{I} )}$ and setting the $\bm{\Phi}$ for the posterior to the identity matrix which results in \eq(\ref{eqn:post-meanField}). The rest of the algorithm stays intact.

%\paragraph{Computational Complexity:}
\textbf{Computational Complexity:} 
To perform the inversion in line 10 of Algorithm~\ref{algo:fixFrmDPP}, we use conjugate gradient(CG) which has the complexity of $O(m \sqrt{k})$, where $k$ is the condition number and $m$ is the number of non-zero entries. We initialize the solver with warm initialization $\bm{\theta}^{t-1}$ which helps greatly (in our experience, CG converges very quickly). 
We currently rely on a MATLAB implementation to prove the concept; a low-rank approximation of $\mathbf{C}$ (similar to LBFGS method) should alleviate the memory complexity. If $\bm{\Phi}\bm{\Phi}'$ is low-rank (which is the case in this paper) the complexity of sampling from the DPP is reduced to $O(d^2 M)$ per iteration where $d$ is the rank of the similarity matrix and $M$ is the number of elements. Computing the marginal likelihood for regression has a closed form that involves inversion of a matrix ($O(J^3)$ where $J$ is the number of selected elements in each iteration). For classification, we use expectation propagation to approximate the marginal likelihood and the complexity of that is defined by $J$ (number of selected elements in each iteration). With smart initialization from the previous iteration, EP converges very quickly. In addition, smart bookkeeping from previous iterations can reduce the number of marginal likelihood computations.

%% file: experiments.tex
\section{Experiments}
\label{sec:experiment}

We show the results for two experiments covering both classification (Section \ref{sec:genNet}) and regression (Section \ref{sec:optGrid}); more experiments are provided in the supplementary material. While in Section \ref{sec:optGrid}, the features themselves are used to define diversity ($\bm{\Phi} = \mathbf{X}$) to penalize collinearity, in Section \ref{sec:genNet} we define the diversity through side information ($\bm{\Phi} \neq \mathbf{X}$). 
In all of our experiments, we fit the parameters of the posterior DPP and compute the MAP subset, $\mathcal{S}^*$, using \cite{nemhauser1978analysis}. Since $\mathbf{L}$ is low rank and due to the numerical scale of the optimal quality score, $\exp (\bm{\theta})$, the local optimization strategy of \cite{mapDPP} failed, hence we use the greedy approach proposed in \cite{nemhauser1978analysis} to approximate the MAP. We compared our models to six other baseline methods: orthogonal matching pursuit (\texttt{OMP}), generalized linear model (GLM) Lasso (\texttt{Lasso}), GLM elastic net (\texttt{eNet}), forward selection (\texttt{FS}), spike-and-slab (\texttt{SpikeSlab}) \cite{carbonetto2012scalable}, and using DPP as prior with a fully factorized mean field (\texttt{DPP-Bernoulli}). \texttt{Lasso} and \texttt{FS} are standard approaches for feature selection using convex and greedy optimization. Elastic net was chosen since the extra $\ell_2$ norm better copes with collinearity better. \texttt{OMP} was chosen since the orthogonality procedure induces some notion of diversity for $\mathbf{\Phi}=\mathbf{X}$. \texttt{SpikeSlab} and \texttt{DPP-Bernoulli} assume Bernoulli and DPP priors for the inclusion of the features respectively but both deploy mean field to approximate the posterior. Parameters of the methods are adjusted to match the number of selected features with the cardinality of $\mathcal{S}^*$.

\subsection{Breast cancer prognosis prediction}
\label{sec:genNet}

\begin{figure*}[tb]
     \centering
     \subfloat[][]
     {
	\includegraphics[height=3.5cm]{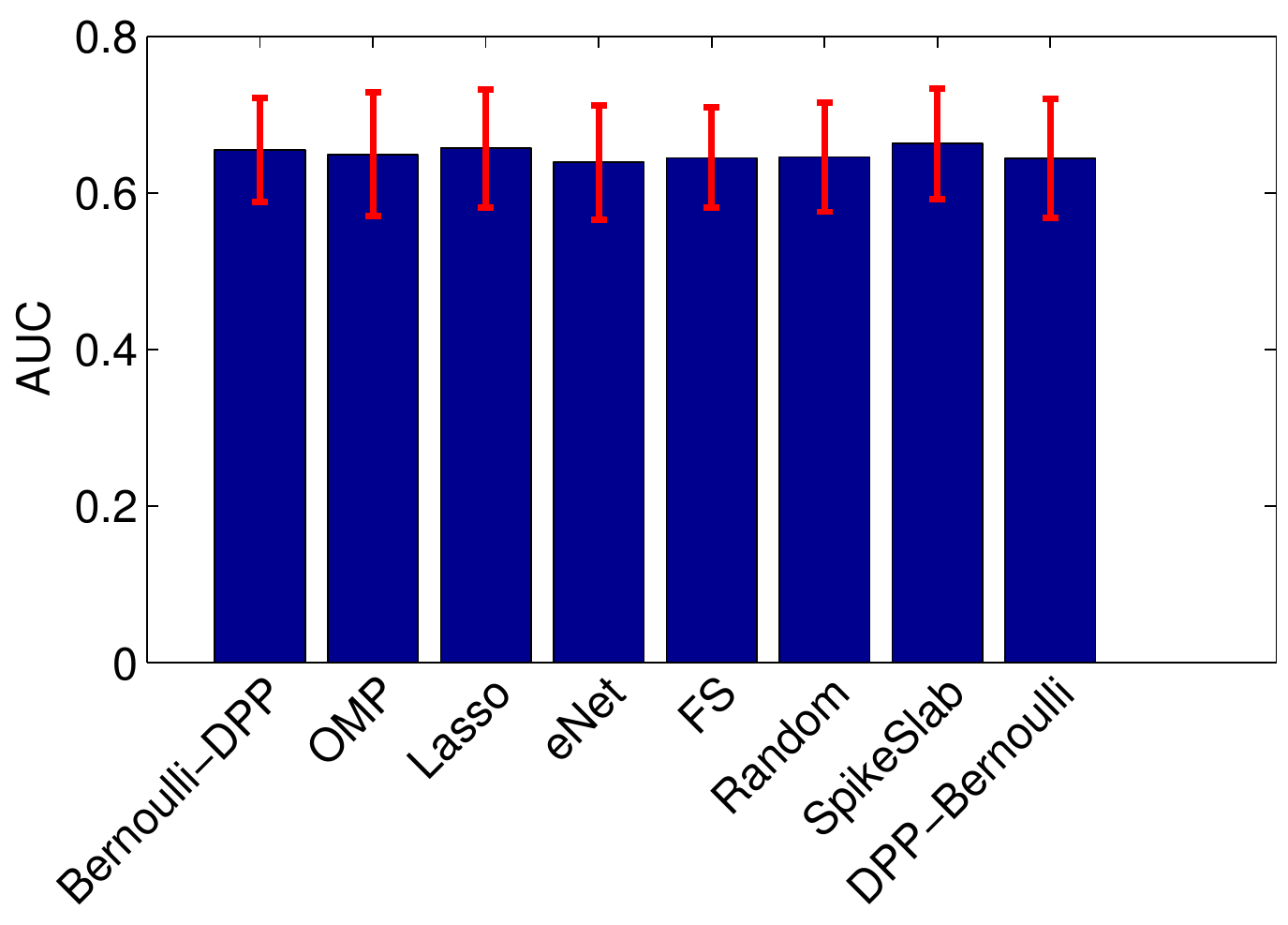}
	\label{fig:genNwk-auc}
     }
     \subfloat[][]
     {
	\includegraphics[height=3.5cm]{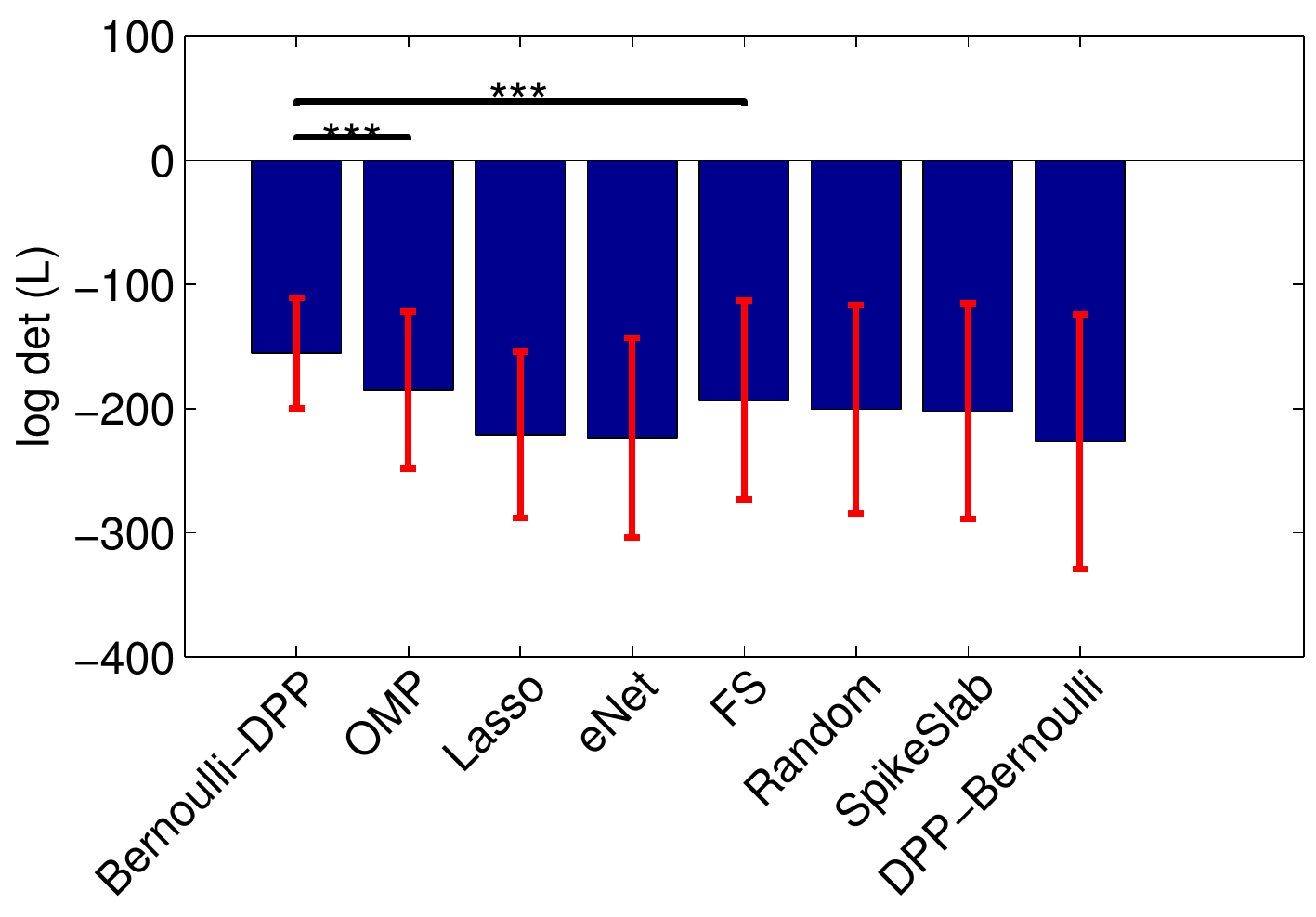}
	\label{fig:genNwk-logDet}
     } 
     \subfloat[][]
     {
	\includegraphics[height=3.5cm]{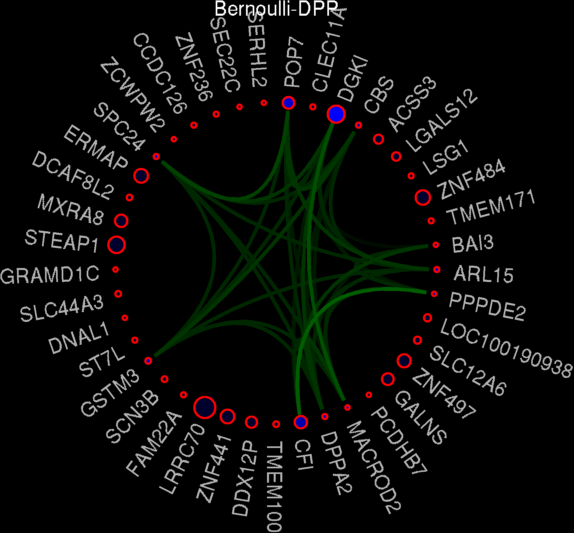}
	\label{fig:genNwk-DPP}
     } 
%      \subfloat[][]
%      {
% 	%\includegraphics[height=3.25cm]{drawings/geneNetwork/Lassov2.pdf}
% 	\includegraphics[scale=0.20]{drawings/geneNetwork/Lasso_final.pdf}
% 	\label{fig:genNwk-Lasso}
%      }\\
     \subfloat[][]
     {
	\includegraphics[height=3.5cm]{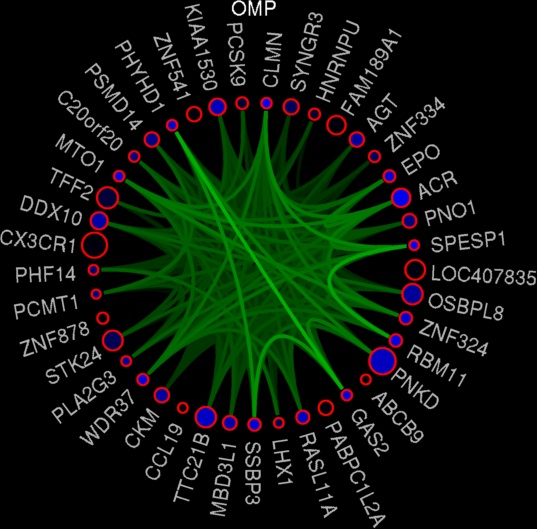}
	\label{fig:genNwk-OMP}
     } 
%      \subfloat[][]
%      {
% 	%\includegraphics[height=3.25cm]{drawings/geneNetwork/Lassov2.pdf}
% 	\includegraphics[scale=0.20]{drawings/geneNetwork/FS_final.pdf}
% 	\label{fig:genNwk-FS}
%      }
%      \subfloat[][]
%      {
% 	%\includegraphics[height=3.25cm]{drawings/geneNetwork/Lassov2.pdf}
% 	\includegraphics[scale=0.20]{drawings/geneNetwork/eNet_final.pdf}
% 	\label{fig:genNwk-eNet}
%      }
     \caption{ \protect\subref{fig:genNwk-auc} Area under curve (AUC) performance averaged over 100 train/test repeats of classification of tumor stages for different methods. \protect\subref{fig:genNwk-logDet} Diversity of selected features, quantified as the determinant of $\mathbf{L}_{\mathcal{S}}$ where $\mathcal{S}$ is the subset of genes selected by each method (higher values indicate more diversity). DPP yields more diverse subset without compromising  accuracy. The asterisks in \protect\subref{fig:genNwk-logDet} indicate statistical significance (based on $p$-value) using a Wilcox rank sum test. 
     %\protect\subref{fig:genNwk-DPP} A connectivity network of all genes selected in more than $10\%$ of the cross validations  for DPP, where an edge between genes $i$ and $j$ indicates that $L_{ij}$ is among the largest $30\%$ of entries of $\mathbf{L}$. \protect\subref{fig:genNwk-OMP} the same for OMP.
     \protect\subref{fig:genNwk-DPP}, \protect\subref{fig:genNwk-OMP} Networks for top 40 genes for \texttt{Bernoulli-DPP} and \texttt{OMP} respectively. The genes are sorted according to the number of times they present in the optimal set in 100 repeats. The radius is proportional to number of times the gene is selected while the color indicates the sum of $L_{ij}$ in that gene.   
    }
     \label{fig:genNwk}
\end{figure*}

In this section, we turn to the motivating application of our method -- finding a diverse set of genes (features) that distinguish stage I and II breast tumors. Constructing accurate classifiers will help identify important biomarkers of breast cancer progression, and furthermore, increasing the functional diversity of selected genes will more likely identify a comprehensive set of cancer-related pathways.  
%We use auxiliary information to define similarities between features that is different from inner product of the features ($\mathbf{X} \neq \boldPhi$).
The main idea is as follows.  Gene expression profiles are the most readily available data for predicting breast cancer prognosis.  However, correlation of gene expression is a relatively poor predictor of correlation in gene function \cite{ppicxcomparison}, and so $\mathbf{X}$ is a poor feature to define functional similarity of genes. Distance between gene pairs in a protein-protein interaction (PPI) network is predictive of gene function \cite{ppicxcomparison}: PPI networks tend to form communities, and genes belonging to the same community perform similar functions.  However, since community detection is very challenging \cite{fortunato2010community}, using network distance to define similarity for DPP avoids a community detection step.    

We first collected 668 subjects from The Cancer Genome Atlas \cite{tcga2012data} with stage I and II breast cancer. We computed normalized expression levels for 13,876 genes for which at least one physical protein-protein interaction was found in the BioGRID database \cite{biogriddata2006}, then focused on the top 2,000 genes with smallest $p$-value (according to a likelihood ratio test for a univariate logistic regression model) with respect to the tumor stages. We then used the BioGRID gene interaction network to compute pairwise similarities between genes (features) as follows: Given the scale-free nature of the network, we first identified hubs of the network as those nodes with total degree higher than 100. For each gene $i$, we then defined its network profile $\mathbf{f}_i$ as a 300-dimensional vector, where each component specifies the shortest path from that gene to a hub. Our feature similarity matrix, $\mathbf{L} = \boldPhi \boldPhi^T$, measures similarity between genes $i$ and $j$ as $L_{ij} = \exp {\left(-||\mathbf{f}_i - \mathbf{f}_j||^2 / \sigma^2 \right)}$ where $\sigma$ is set to $3$, approximately the average pairwise distance between genes. 

\figurename{ \ref{fig:genNwk-auc}} and \figurename{ \ref{fig:genNwk-logDet}} show that \texttt{Bernoulli-DPP} identifies gene sets significantly more diverse than all other methods, without compromising prediction accuracy. We note that imposing DPP as an approximate posterior leads to more diverse set than having DPP as prior in \texttt{DPP-Bernoulli}. We also randomly select genes with low $p$-value to see if it leads to diverse set (\ie \texttt{Random} in \figurename{\ref{fig:genNwk-logDet}}). 
Although AUC of Random is in the same range as the other method, the diversity is below \texttt{Bernoulli-DPP} and \texttt{FS}. \texttt{SpikeSlab} produces good accuracy but the selected genes are basically top genes according to $p$-value (not shown in the figure) and the gene set is not diverse. 

We next assessed whether the diverse set of genes identified by \texttt{Bernoulli-DPP} pinpoints pathways involved in breast cancer. We first divided the 2,000 genes into 410 communities \cite{linkcomm2010} using the BioGRID network.  Based on 20 different cross-validation runs using all five methods, we identified 18 communities within the network that were selected more often (in at least 20\% of the runs) by the \texttt{Bernoulli-DPP} than any other method.  We found these DPP-preferred communities were enriched in genes related to cell cycle checkpoints, metabolism, DNA repair, predicted breast cancer gene modules, and interactors of several known cancer genes such as BRCA2, AATF, ANP32B, HDAC1, and PRKDC, among others \cite{msigdb05}. We also observed enrichment in genes down-regulated in activated T-cells relative to naive T-cells and other immune cell types. Although the role of T-cells in tumor immunity is not fully understood, recent work has implicated immune cell activity (and T-cell infiltration in general) with breast cancer survival \cite{gutrantien13,peguillet14} and our results both support these findings and potentially widen the set of genes that may need to be investigated further for anti-tumor properties.

\subsection{Optimal Gridding of Spatial Process Convolution Models}
\label{sec:optGrid}

\begin{figure*}[tb]
	\centering
	\subfloat[][]
	{
		%\begin{minipage}{5.0cm}
		\includegraphics[width=5.3cm]{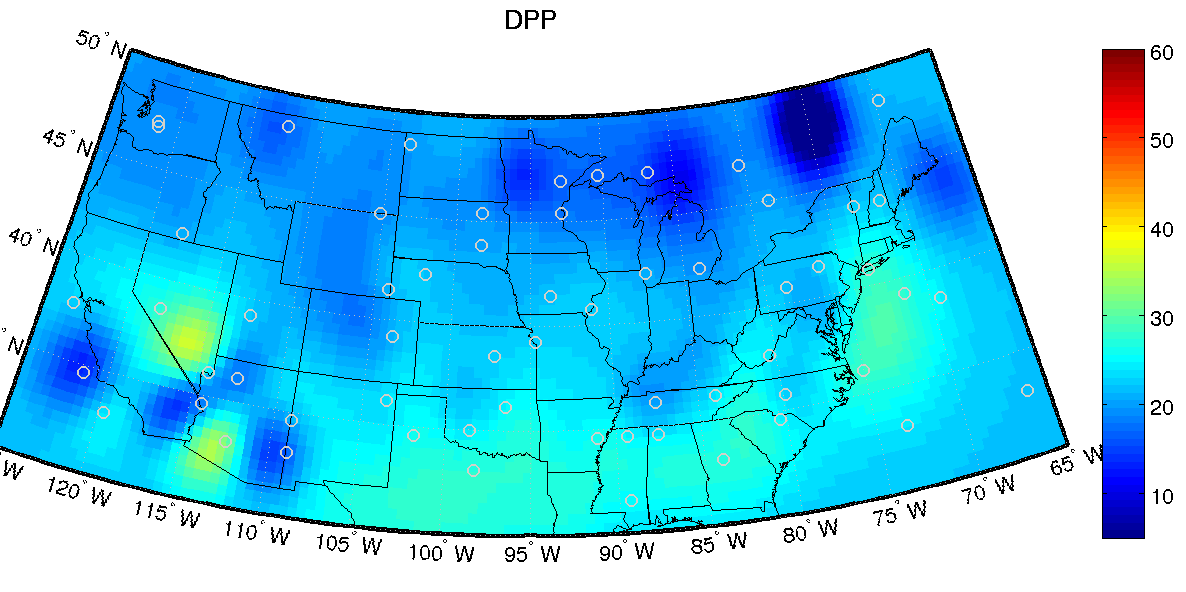}
		%\end{minipage} 
		\label{fig:tempDPP}
	}
	\subfloat[][]
	{  
		%\begin{minipage}{3.25in}
		\includegraphics[width=5.3cm]{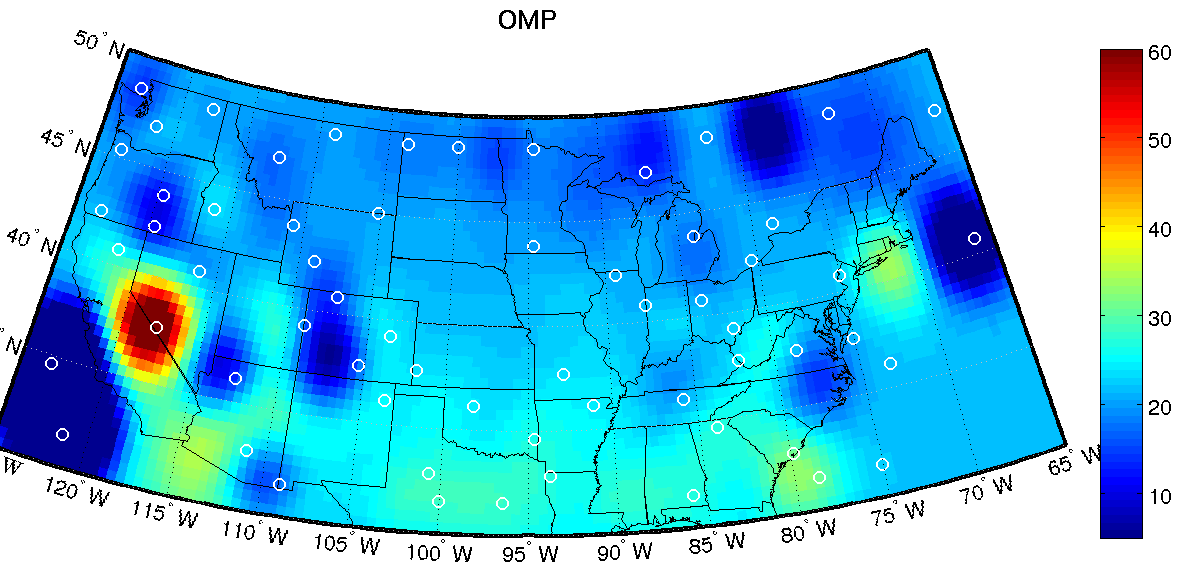}
		%\end{minipage}
		\label{fig:tmpOMP}
	} \\
	\subfloat[][]
	{  
		\includegraphics[height=3.7cm]{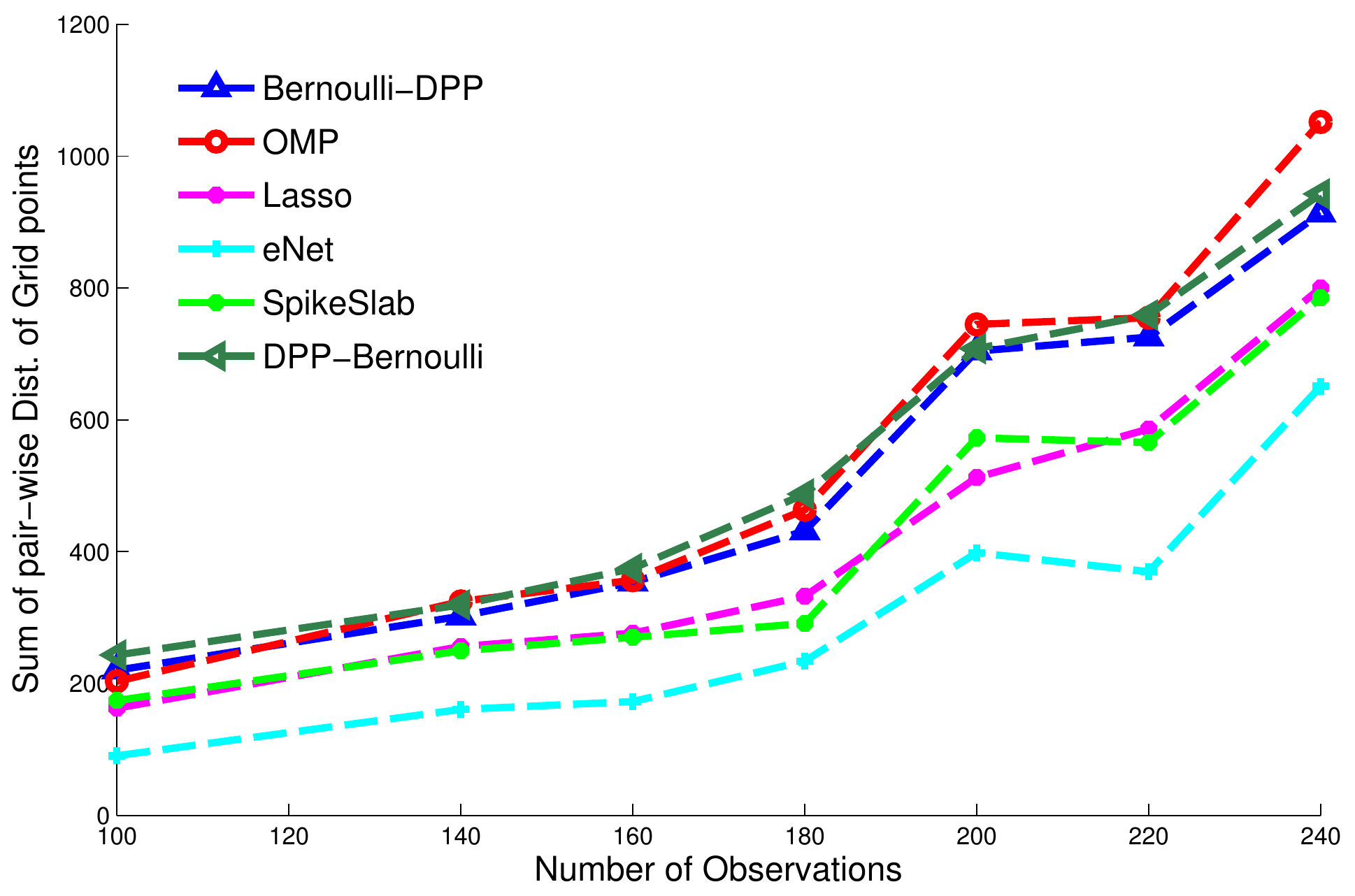}
	        \label{fig:USTemp-pdist}
	} 
	\subfloat[][]
	{  
		\includegraphics[height=3.7cm]{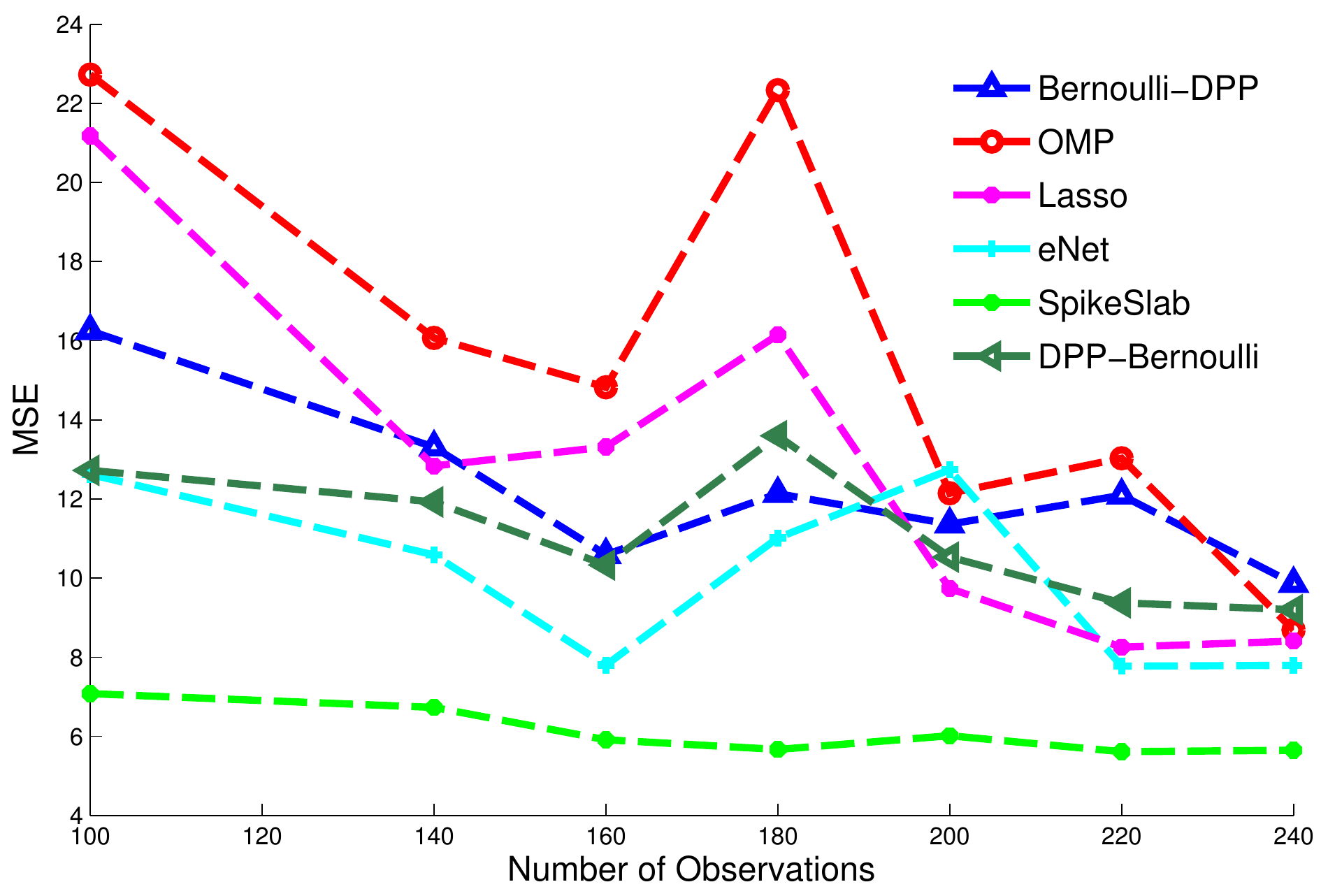}
	        \label{fig:USTemp-mse}
	}
	\caption{\protect\subref{fig:tempDPP} and \protect\subref{fig:tmpOMP} show examples of gridding (white circles) and the prediction for \texttt{Bernoulli-DPP} and \texttt{OMP}. The grid points are spread out in both methods but \texttt{OMP} suffers from overshooting (or undershooting) prediction (\eg over California in \protect\subref{fig:tmpOMP}). \protect\subref{fig:USTemp-pdist}  and \protect\subref{fig:USTemp-mse}  show the average of pairwise distance  between selected grid points and MSE respectively for different number of measurements. Diversity promoting methods (\ie \texttt{Bernoulli-DPP}, \texttt{DPP-Bernoulli}, \texttt{OMP}) performs similarly in term of diversity while DPP-related method are slightly better in term of MSE. \texttt{SpikeSlab} and \texttt{eNet}  outperform other methods in term of MSE but as shown in \protect\subref{fig:USTemp-pdist} the selected grid points are much closer to each other. \texttt{OMP} and DPP-related methods seem to have a better balance between MSE and having distributed grids particularly for a large number of measurements.}
	\label{fig:USJulyTemp}
\end{figure*}

We now demonstrate the method applied to a problem in spatial statistics, namely constructing a non-stationary Gaussian process (GP) model in a computationally efficient way. One way to construct a GP is to convolve white noise $x(s)$ by a continuous function: $z(s) = k(s) * x(s)$ ($s \in \Omega \subset \mathbb{R}^d$). The resulting GP has covariance $\int_{\Omega}{ k(u-d)k(u) du }$. Higdon \etal \citep{higdon1998process} suggested to define $z(s)$ to be zero mean GP and instead of defining the covariance directly, determine it implicitly through a latent white noise process $x(s)$ and smoothing kernel $k(s)$. $x(s)$ are restricted to be nonzero at the spatial sites $\omega_1,\dots,\omega_M \in \Omega$ and each is drawn from $\mathcal{N}(\cdot;0,\sigma_x^2)$. The resulting GP is $z(s) = \sum_{j=1}^{M}{x_j k(s- \omega_m )}$. One can view $\omega_m$ as (irregularly spaced) grid points. Assuming $z$ is observed with some noise, the problem is equivalent to 
regression, where feature selection is equivalent to finding the optimal locations for the spatial bases.

Our objective is to find the optimal gridding of spatial domain for prediction, while also ensuring a broad spread across the spatial domain. Each grid point covers an areas but here in a 2-D domain.  The grid points are the centers of the basis vectors which are isotropic Gaussian bumps on three different scales. Notice that having spatially spread out basis functions boils down to having basis functions with little overlap. Hence diversity may simply be computed as the inner product between basis vectors, i.e. $\bm{\Phi}=\mathbf{X}$.  For this experiment, the temperature is measured for the month of July at $476$ sensors located across the United States. We randomly choose varying number of sensors as a training set and evaluate the performance on the left out sensors for all methods. This procedure was repeated $20$ times and we report the average performance (MSE) in \figurename{ \ref{fig:USTemp-mse}}. \figurename{ \ref{fig:USTemp-pdist}} reports the average pairwise distance between the selected points. In term of MSE, both \texttt{DPP-Bernoulli} and \texttt{Bernoulli-DPP} perform better than \texttt{OMP} and slightly better than \texttt{Lasso} but \texttt{SpikeSlab} outperforms the other methods. However, as is evident in \figurename{ \subref{fig:USTemp-pdist}}, \texttt{SpikeSlab} does not produce spread out grid points which was the main objective. In contrast, \texttt{DPP-Bernoulli}, \texttt{Bernoulli-DPP} and \texttt{OMP} strike a good balance between prediction and diversity. Examples of the reconstructions are shown in \figurename{ \ref{fig:tempDPP}} for \texttt{Bernoulli-DPP} and  \figurename{\ref{fig:tmpOMP}} for \texttt{OMP}. \texttt{OMP} tends to overshoot in areas with few measurements -- a trait also observed in the simulation (see the supplementary material). It is also interesting to see that when $\bm{\Phi} = \mathbf{X}$ having DPP as prior or approximate posterior perform similarly.

%% file: conclusion.tex
\section{Conclusions}

In this paper, we have proposed a probabilistic method for diverse feature selection. 
We proposed to approximate the posterior distribution with DPPs as a computationally elegant way to encourage diversity. Our approach selects the most representative items in communities of relevant items. Similarity between items can be encoded through the inner product between features to discourage collinearity (similar to OMP) or may be defined based on side information (\eg Section \ref{sec:genNet}). Our model therefore allows features and similarities to be different ($\boldPhi \neq \mathbf{X}$). When $\boldPhi = \mathbf{X}$, in the experiments in Section \ref{sec:optGrid}, using DPP as an approximate posterior performs similarly to using DPP as prior with mean-field approximation.

While learning the parameters of DPPs is an active research area, we have shown a computationally efficient strategy for learning the parameters in our variational approach. As far as we know, our method is the first variational method used to learn the parameters of a DPP distribution. 

Our algorithm relies on sampling from the DPP, which involves a singular value decomposition (SVD) in each iteration. SVD is not very stable for matrices with very large condition numbers, hence it would be interesting to explore other parametrization of DPPs, such as those in \cite{affandi_learningDPP}. An alternative parametrization can hopefully improve the condition number of the optimal kernel matrix $\mathbf{L}$ and improve the performance of the MAP approximation \cite{mapDPP}. 

In conclusion, imposing DPP as an approximate posterior selects more diverse features without compromising the accuracy; further, it allows for sampling-based quantification of uncertainty. If the posterior distribution is multi-modal, sampling from the model can provide an alternative solution - something not possible with OMP. In fact, the simulated examples demonstrated that DPP is more robust than OMP (see supplement). %The proposed method not only allows for probabilistic measures of uncertainty, but it also has similar prediction performance as competing methods while yielding feature sets which are considerably more diverse.

%% file: appendix.tex
\section{Algorithm for Posterior Approximation}
\label{sec:appendix}
The Algorithm \ref{algo:fixFrmDPP} can be used to approximate the posterior for both \texttt{DPP-Bernoulli} and \texttt{Bernoulli-DPP}. For \texttt{Bernoulli-DPP}, we set $p(\boldGamma) = \prod_{m=1}^{M}{\alpha^{\gamma_m} (1-\alpha)^{1-\gamma_m}}$. 

In case of \texttt{DPP-Bernoulli}, DPP and Bernoulli (\ie fully factorized posterior) are deployed as the prior and the approximate posterior respectively. We just need to modify $p(\bm{\gamma}) = \frac{\det ( [\mathbf{L}]_{\bm{\gamma}} )}{\det ( \mathbf{L} + \mathbf{I} )}$ and $\bm{\Phi}=\mathbf{I}$.  

\begin{algorithm}[h]
\KwIn{ Similarity features $\bm{\Phi}$, a function to compute/approximate the restricted marginal likelihood $p(y | \boldGamma)$, $p(\boldGamma)$, initial cardinality of DPP $\kappa$ , number of iterations: $N$. }
\KwOut{Parameters of the posterior ($q$): $\boldTheta$}

Adjust expected cardinality of the DPP by solving for $\theta_0$ in $ \sum_{i} \frac{e^{\theta_0} \lambda_i }{1 + \lambda_i e^{\theta_0}} = \kappa $ \;

Initialize $\boldTheta = \theta_0 \mathbbm{l}$, $\mathbf{L} =  e^{\theta_0/2} \bm{\Phi} \bm{\Phi}^T e^{\theta_0/2}$, and $\mathbf{K} = \mathbf{L} (\mathbf{L} + \mathbf{I})^{-1}$ \;

Set $(\mathbf{C}_1)_{ii} = K_{ii} $, $\mathbf{g}_1 = \mathbf{C}_1 \tilde{\boldTheta} $, and $\bar{\mathbf{C}}, \bar{\mathbf{g}} = 0$ \;

%$w = \frac{1}{\sqrt{N}}$ \;

\For{$t \gets 1$ \textbf{to} $N$}{
    Draw a set  from current posterior approximation DPP: $\boldGamma^{*}_{t} \sim q_{\tilde{\boldTheta}_{t}}$ \;
      
    Set $ \hat{\mathbf{g}}_t = \tilde{\boldGamma^{*}_{t}} \log p(y|\boldGamma^{*}_{t})p(\boldGamma^{*}_{t})$ \;

    Set $ \hat{\mathbf{C}}_t = \tilde{\boldGamma^{*}_{t}} \tilde{\boldGamma^{*}_{t}}^{T}$ or current estimate of $\mathbf{K}_{\mathbf{\theta}_{t}}$\;

   Set $\mathbf{g}_{t+1} = (1-w) \mathbf{g}_t + w \hat{\mathbf{g}}_t $ \;
   %Set $\mathbf{g}_{t + 1} = \left( 1 - w \right) \mathbf{g}_t + w \tilde{ \boldGamma_t } \log p \left( \boldGamma_t, y \right) $ \;
  
    Set $\mathbf{C}_{t+1} = (1-w) \mathbf{C}_t + w \hat{\mathbf{C}}_t$  \;
    %Set $\mathbf{C}_{t + 1} = \left( 1 - w \right) \mathbf{C}_t + w \tilde{\boldGamma_t }  \tilde{ \boldGamma_t }^T$  \;

    Solve $\tilde{\boldTheta}_{t+1} = \mathbf{C}_{t+1}^{-1} \mathbf{g}_{t+1}$ \;

    \If {$t > N/2$ }{
      Set $\bar{\mathbf{g}} = \bar{\mathbf{g}} + \hat{\mathbf{g}}_t $\;
      Set $\bar{\mathbf{C}} = \bar{\mathbf{C}} + \hat{\mathbf{C}}_t $\;
    }
}

\Return{$\boldTheta = \bar{\mathbf{C}}^{-1} \bar{\mathbf{g}}$}\;
\caption{Variational Learning for Diverse Variable Selection}
\label{algo:fixFrmDPP}
\end{algorithm}